\pdfoutput=1

\documentclass[11pt]{article}

\pdfoutput=1
\usepackage[T1]{fontenc}
\usepackage[utf8]{inputenc}
\usepackage{textcomp}

\DeclareUnicodeCharacter{03BA}{\ensuremath{\kappa}} 
\DeclareUnicodeCharacter{2013}{--}   
\DeclareUnicodeCharacter{2212}{-}    
\DeclareUnicodeCharacter{00A0}{~}    

\usepackage{EMNLP2023}

\usepackage{times}
\usepackage{latexsym}
\usepackage{parskip}  

\usepackage[T1]{fontenc}
\usepackage{amsmath}

\usepackage[utf8]{inputenc}

\usepackage{microtype}

\usepackage{inconsolata}

\title{Beyond the Score: Uncertainty-Calibrated LLMs \\for Automated Essay Assessment}

\author{
Ahmed Karim \\
King's College London \\
{\small\texttt{ahmed.karim@kcl.ac.uk}}  
\And
Qiao (Judy) Wang \\
Hosei University \\
{\small\texttt{judy.wang@hosei.ac.jp}}
\And
Zheng Yuan \\
University of Sheffield \\
{\small\texttt{zheng.yuan1@sheffield.ac.uk}}
}

\begin{document}
\maketitle

\begin{abstract}
Automated Essay Scoring (AES) systems now attain near–human agreement on some public benchmarks, yet real-world adoption—especially in high-stakes examinations—remains limited.  A principal obstacle is that most models output a \emph{single} score without any accompanying measure of confidence or explanation.  We address this gap with \emph{conformal prediction}, a distribution-free wrapper that equips any classifier with set-valued outputs enjoying formal coverage guarantees.  Two open-source Large Language Models—Llama-3 8B and Qwen-2.5 3B—are fine-tuned on three diverse corpora (ASAP, TOEFL11, Cambridge-FCE) and calibrated at a 90\% risk level.  Reliability is assessed with \textbf{UAcc}, an uncertainty-aware accuracy that rewards models for being both correct and concise.  To our knowledge, this is the first work to combine conformal prediction and UAcc for essay scoring.  The calibrated models consistently meet the coverage target while keeping prediction sets compact, indicating that open-source, mid-sized LLMs can already support teacher-in-the-loop AES; we discuss scaling and broader user studies as future work.
\end{abstract}

\section{Introduction}

Automated Essay Scoring (AES) has evolved rapidly—from linear regressors built on handcrafted features \citep{phandi-etal-2015-flexible}, through CNN–LSTM hybrids that capture local and long-range coherence \citep{taghipour-ng-2016-neural}, to transformer encoders such as R2BERT that pair BERT representations with joint regression–ranking objectives and reach state-of-the-art agreement on ASAP essays \citep{yang-etal-2020-enhancing}.  The latest step is the move to (open-source) Large Language Models (LLMs), e.g., fine-tuned Llama variants now approach human-human agreement on several AES benchmarks \citep{10.1145/3706468.3706507}.

Headline accuracy, however, is not enough for high-stakes settings such as \textsc{TOEFL}\footnote{\url{https://www.ets.org/toefl.html}} or \textsc{IELTS}\footnote{\url{https://ielts.org/}}, where a single mis-scored script can determine admission or visa status.  Exam boards need calibrated confidence.  Common approaches to measuring uncertainty include Monte-Carlo dropout \citep{pmlr-v48-gal16}, deep ensembles \citep{10.5555/3295222.3295387} and Bayesian neural networks \citep{unknown}. These methods are effective but either multiply inference cost or offer no finite-sample guarantees.

Conformal prediction (CP) \citep{DBLP:journals/corr/abs-2107-07511} provides such guarantees by wrapping any classifier with a \emph{set-valued} output that contains the true label with user-chosen probability.  CP has improved reliability in tasks from surrogate models \citep{surrogate} to question answering, yet it has not been applied to AES, and no study has linked calibration quality to scoring usefulness.  We bridge that gap with \textbf{UAcc}—an uncertainty-aware accuracy that rewards models that are correct and selective \citep{ye2024llm_uq}.

In this paper, we fine-tune two state-of-the-art LLMs (Llama-3 8B \citep{DBLP:journals/corr/abs-2407-21783} and Qwen-2.5 3B \citep{qwen}) on three public AES benchmarks (ASAP \citep{kaggleASAP}, TOEFL11 \citep{ldc2014t06}, Cambridge-FCE \citep{yannakoudakis-etal-2011-new}).  Each scorer is then calibrated with conformal prediction so that its prediction set is guaranteed, by construction, to contain the true score in at least 90 \% of future essays.  We evaluated these calibrated models with the uncertainty-aware accuracy \textbf{UAcc}, alongside standard accuracy and quadratic-weighted $\kappa$.  Across all corpora, the models meet the 90\% coverage guarantee while keeping prediction sets tight, showing that uncertainty-aware AES is already practical with mid-sized, openly licensed LLMs.  By uniting modern language models, distribution-free calibration, and an uncertainty-sensitive metric, we provide the first comprehensive picture of trustworthy essay scoring across multiple proficiency tests.

\section{Background}
\label{sec:background}

This work combines a standard essay-scoring model with \emph{conformal prediction} so that every prediction comes with a statistically valid notion of confidence.

\subsection{Essay-scoring task}
An essay \(x\) must receive one label \(y\) from a fixed set of \(K\) possible scores (e.g.\ the integers 2–12 for ASAP, or the three bands \textit{low/med/high} for TOEFL).  
A neural scorer \(f\) takes the essay text and outputs a probability for each label; we denote that distribution by \(\hat{p}(y\mid x)\).

\subsection{Conformal prediction (CP)}
Conformal prediction turns those probabilities into a \textbf{prediction set}
\(C_{\alpha}(x) \subseteq \{1,\dots,K\}\) that is guaranteed to contain the true score with high probability.  
Formally, for a user-chosen risk level \(\alpha\) (we use \(\alpha=0.1\)), CP ensures
\begin{equation}
\Pr\!\bigl[y \in C_{\alpha}(x)\bigr] \;\ge\; 1-\alpha
\label{eq:coverage}
\end{equation}
so the true score falls outside the set in at most 10\% of future essays.

\paragraph{How conformal sets are constructed.}
To build a prediction set using conformal prediction, the data is first split into three parts: a training set for fitting the model, a calibration set for estimating uncertainty, and a test set for evaluation.

For a given model \(f\) and input essay \(x\), the score assigned to each possible label \(y\) is defined as
\begin{equation}
s(x, y) = 1 - \hat{p}(y \mid x)
\label{eq:score}
\end{equation}
where \(\hat{p}(y \mid x)\) is the model’s predicted probability. This is known as the least-ambiguous classifier (LAC) score—lower scores indicate higher confidence.

Using the calibration set, the conformal algorithm computes a threshold \(q_\alpha\) such that at most an \(\alpha\) fraction of calibration scores exceed it. Then, for any new essay \(x\), the prediction set is formed by including all labels with scores below this threshold:
\begin{equation}
C_\alpha(x) = \{ y \in \mathcal{Y} \mid s(x, y) \le q_\alpha \}
\label{eq:prediction-set}
\end{equation}

This guarantees that the prediction set contains the true label with probability at least \(1 - \alpha\).

\subsection{Metrics}
We evaluate models using both standard and uncertainty-aware criteria. Quadratic-weighted \(\kappa\) (QWK) measures agreement with human raters while penalising larger score discrepancies more heavily than near misses, which is natural for ordinal rubrics. In addition to accuracy and QWK, we report three key metrics specific to conformal prediction:  
(i)~\textbf{Coverage}, the proportion of test essays for which the true label is contained in the prediction set \(C_{\alpha}(x)\);  
(ii)~\textbf{Average set size}, measuring how many labels are typically included—smaller is better;  
and (iii)~\textbf{UAcc} which balances correctness and conciseness via
\begin{equation}
\text{UAcc} = \text{Accuracy} \times \sqrt{\frac{K}{\text{avg.}\,|C_{\alpha}(x)|}}
\label{eq:uacc}
\end{equation}

where \(K\) is the number of classes.  
UAcc penalises large or overly cautious prediction sets, rewarding models that are both accurate and selective.

\begin{table*}[t]
\centering
\setlength{\tabcolsep}{6pt}
\begin{tabular*}{\textwidth}{@{\extracolsep{\fill}} l r r r l}
\hline
Corpus & Train & Cal & Test & Labels \\
\hline
ASAP P1            & 1\,248 &   268 &   267 & 11-way (2–12) \\
TOEFL11            & 8\,470 & 1\,815 & 1\,815 & low / med / high \\
Cambridge-FCE      & 1\,742 &   373 &   373 & low / med / high \\
\hline
\end{tabular*}
\caption{Dataset sizes after a 70 / 15 / 15 train–calibration–test split. These are the number of essays in each split}
\label{tab:data}
\end{table*}

\section{Experimental Setup}
\label{sec:setup}

\subsection{Models and tokenisation}
We experiment with two openly licensed generative LLMs: \textbf{Llama-3 8B} and \textbf{Qwen-2.5 3B}.  
Both are loaded via HuggingFace Transformers with 4-bit quantisation; special tokens and maximum context length follow the model cards.  
For each corpus we prepend a short instruction—\textit{“Read the essay and output a single score:”}—and rely on the tokenizer to convert either the integer label (ASAP) or the band token \emph{low/medium/high} (TOEFL11, FCE) into a single ID, so that the final token distribution can be treated as a 3- or 11-way classifier without adding new parameters.

\subsection{Fine-tuning}
Training is performed on a single Nvidia A100-40GB GPU for eight epochs with AdamW (learning rate $1\times10^{-5}$). We use a global batch size of 8 and a fixed random seed (42) to ensure reproducibility.

\subsection{Calibration and test split}
After fine-tuning, the original validation\,+\,test portion of each corpus is split once into equal-sized \textit{calibration} and test sets (15 \% / 15 \% of the full data; exact counts in Table \ref{tab:data}).  
Calibration essays never influence model weights.

\subsection{Conformal prediction}
For every essay–label pair we compute the least-ambiguous classifier score
\(s(x,y)=1-\hat{p}(y\mid x)\), where \(\hat{p}\) is the model’s softmax probability.  
The $(1-\alpha)$ quantile of these scores on the calibration set with $\alpha = 0.1$ yields the threshold \(q_{\alpha}\).  
At inference time we return all labels whose scores fall below \(q_{\alpha}\), guaranteeing that the prediction set contains the true label in at least 90 \% of future essays.

\subsection{Evaluation metrics}
We report conventional accuracy and QWK together with three uncertainty-aware measures introduced in Section \ref{sec:background}:  
\textbf{Coverage}, the empirical proportion of essays whose true label lies in the prediction set; \textbf{Average set size}, a proxy for informativeness; and \textbf{UAcc}, which trades off accuracy against set size.

\subsection{Datasets}
\label{sec:data}

Table~\ref{tab:data} lists the three corpora and the statistics derived from our 70 / 15 / 15 train–calibration–test split.

\paragraph{ASAP Prompt 1}
Essays written by secondary-school students and graded on an eleven-point scale (2–12).  
We keep the full scale, as retaining an intermediate label space allows us to study how uncertainty behaves when the number of possible scores increases—something neither TOEFL11 (3 classes) nor FCE (3 bands) can reveal.

\paragraph{TOEFL11}
Internet-based TOEFL essays pre-labelled \emph{low}, \emph{medium} or \emph{high}.

\paragraph{Cambridge-FCE}
Scripts scored holistically 1–40.  To align with TOEFL11 and keep prediction sets interpretable, we divide the range into equal-width thirds—1–18 (\emph{low}), 19–30 (\emph{medium}) and 31–40 (\emph{high}).  This heuristic balances the three classes and prevents prediction sets from ballooning to forty labels; exploring finer buckets is left for future work.

\section{Results and Discussion}
\label{sec:results}

\begin{table*}[t]
\centering
\setlength{\tabcolsep}{7pt}
\begin{tabular}{l l c c c c c c}
\hline
Dataset & Model & QWK & Acc. & F1 & Coverage & Avg.\ $|C|$ & UAcc \\
\hline
ASAP P1 & Qwen-2.5 3B & 0.69 & 0.50 & 0.45 & 0.91 & 3.51 & 0.88 \\
        & Llama-2 7B  & \textbf{0.82} & \textbf{0.54} & \textbf{0.52} & \textbf{0.91} & \textbf{2.74} & \textbf{1.08} \\
        & Llama-3 8B  & 0.80 & \textbf{0.54} & 0.51 & \textbf{0.91} & 2.81 & 1.07 \\
\hline
TOEFL11 & Qwen-2.5 3B & 0.69 & 0.77 & 0.76 & 0.89 & 1.32 & 1.16 \\
        & Llama-3 8B  & \textbf{0.70} & 0.77 & \textbf{0.77} & 0.89 & \textbf{1.29} & \textbf{1.17} \\
\hline
Cambridge-FCE & Qwen-2.5 3B & 0.16 & 0.65 & 0.62 & \textbf{0.95} & 2.30 & 0.74 \\
              & Llama-3 8B & \textbf{0.28} & \textbf{0.66} & \textbf{0.64} & 0.88 & \textbf{1.74} & \textbf{0.87} \\
\hline
\end{tabular}
\caption{Calibrated performance on three corpora.}
\label{tab:calibrated-results}
\end{table*}

\subsection{Calibrated performance and set compactness}
Across all three corpora, the calibrated Llama models achieve the highest QWK, confirming that stronger back-bones still translate into better agreement with human graders even after quantisation and LoRA fine-tuning \citep{hu2022lora}.  Crucially, they do so while returning the \emph{tightest} prediction sets: roughly 2.7 labels on the 11-way ASAP rubric and fewer than two labels on the three-class TOEFL11 and FCE tasks.  Those concise sets lift UAcc above competing systems that share the same point accuracy.  In other words, the Llama scorers are not merely correct; they are confident enough to commit to a smaller subset of possible scores, which reduces the burden on any downstream human reviewer.

\subsection{UAcc and operational impact}
Because UAcc rescales accuracy by $\sqrt{K/|C|}$, a system can gain either by raising raw accuracy or by shrinking its prediction sets.  On ASAP, Llama-2 and Qwen differ by only four accuracy points (0.54 vs 0.50), yet Llama’s sets are 0.8 labels tighter, boosting UAcc from 0.88 to 1.08.  In practice that means nearly 20 \% fewer essays would be flagged for manual review at the same error rate—an operationally significant saving.

\subsection{Accuracy vs.\ F1 under class imbalance}
Accuracy can be misleading whenever the label distribution is skewed. F1 balances precision and recall, revealing whether a model simply exploits the majority label or performs consistently across bands. In our results the gap between accuracy and F1 is small, confirming that the calibrated LLMs do not over-predict a single band; nonetheless, reporting F1 guards against potential imbalance and strengthens the claim that the models generalise across proficiency levels.

\subsection{Coverage and label-space effects}
Empirical coverage lies within one percentage point of the 90\% target on every dataset, demonstrating that a single conformal wrapper generalises from an 11-point rubric (ASAP) to 3-labeled (TOEFL11, FCE) despite the shift in prompt style, score range and proficiency level.  The larger prediction sets observed on ASAP reflect the richer label space: with eleven possible scores the model must sometimes hedge between adjacent grades, a phenomenon less common in the three-band corpora.

\subsection{ASAP Prompt~1 baselines in context}
Table~\ref{tab:asap-baselines} summarises published ASAP Prompt~1 QWK results and situates our numbers alongside prior work.

\subsection{Why FCE QWK is lower than TOEFL11}
The absolute QWK numbers on FCE are markedly lower than on TOEFL11, even though both datasets use the same low/medium/high mapping. Two factors help to explain the gap. First, the FCE essays are mapped \emph{post-hoc} from a 40-point holistic scale, and quadratic κ penalises any band disagreement proportionally to the original distance on that underlying scale; a one-band slip therefore receives a much heavier penalty than in TOEFL11, whose native rubric already has three discrete labels. Second, the FCE corpus is almost one order of magnitude smaller than TOEFL11, magnifying the impact of label noise and leaving less data for both fine-tuning and calibration. Taken together, these artefacts depress QWK even when coverage and UAcc remain competitive.

\subsection{Takeaways}
Overall, these findings show that mid-sized, openly licensed LLMs already deliver high scoring accuracy together with calibrated, interpretable uncertainty which are key prerequisites for deployment in high-stakes assessment. The consistent edge of Llama-3 over its smaller Qwen counterpart confirms that parameter count and pre-training data still matter, yet the margin is small enough to keep lower-footprint models in serious contention wherever hardware or licensing constraints apply. Singleton prediction sets indicate high certainty; larger sets flag scripts for human review. We plan a small teacher-in-the-loop study in follow-up work.

\begin{table*}[t]
\centering
\small
\setlength{\tabcolsep}{5pt}
\begin{tabular*}{\textwidth}{@{\extracolsep{\fill}} l c c}
\hline
Model (paper) & Training epochs / runs & QWK (ASAP Prompt 1) \\
\hline
EASE (SVR) \citep{taghipour-ng-2016-neural}        & —         & 0.781 \\
LSTM (10×) \citep{taghipour-ng-2016-neural}        & 10 runs   & 0.808 \\
Ensemble CNN+LSTM \citep{taghipour-ng-2016-neural} & 10 runs   & 0.821 \\
R2BERT \citep{yang-etal-2020-enhancing}            & 30 epochs & 0.817 \\
Fine-tuned GPT-3.5 \citep{10.1145/3706468.3706507} & 10 epochs & 0.740 \\
Fine-tuned LLaMA-3 (2-pt) \citep{10.1145/3706468.3706507} & 10 epochs & 0.714 \\
\hline
\textbf{Our LLaMA-3 8B} (8 ep) & 8 epochs & \textbf{0.800} \\
\textbf{Our LLaMA-2 7B} (8 ep) & 8 epochs & \textbf{0.823} \\
\hline
\end{tabular*}
\caption{Published ASAP Prompt 1 baselines vs.\ our systems.}
\label{tab:asap-baselines}
\end{table*}

\section{Conclusion}

We set out to answer whether modern large language models can score essays \emph{and} express calibrated uncertainty in a way that is practical for high-stakes assessment.  
By wrapping two LLMs—Llama-3 8B and Qwen-2.5 3B—with conformal prediction and judging them with the uncertainty-aware metric \textbf{UAcc}, we showed that a single, distribution-free calibration step delivers near-perfect coverage (90 \%) across three very different corpora.  
The stronger Llama backbone achieves the best trade-off between agreement with human graders (QWK) and prediction-set tightness, yet the gap to the smaller Qwen model is modest—evidence that trustworthy AES does not require flagship-scale models.  
Taken together, these results provide the first end-to-end demonstration that mid-sized, openly licensed LLMs can power calibrated, human-in-the-loop essay scoring systems today, while laying the groundwork for future studies on model size, finer FCE banding, and rubric-aware prompting.

\textbf{Future work} will probe the trade-off of model size and performance more systematically: we plan to train a spectrum of model sizes (1B to 13B) from several families to quantify when accuracy and UAcc begin to show diminishing returns. On the data side, we will experiment with finer-grained buckets for the FCE corpus and, more generally, with ordinal-aware conformal scores that respect the underlying scale. Finally, we intend to condition prompts on essay characteristics—length, discourse structure to see whether rubric-aware prompting can tighten prediction sets still further without sacrificing coverage.

\section*{Limitations}

This study is confined to English and to three public essay corpora, two of which we deliberately reduce to a three-band rubric for comparability.  Although conformal prediction delivers the promised 90 \% coverage under these conditions, the guarantee relies on calibration and test data being exchangeable; topic drift, candidate demographics or language transfer effects in a real exam session could weaken reliability.  

Our choice of equal-width bands for the 1–40 Cambridge-FCE scale is a heuristic that balances class counts but may mask finer proficiency distinctions.  Likewise, we retain ASAP’s full 11-point rubric to explore class-rich uncertainty, yet that decision limits direct comparison across datasets.

Additionally, our equal-width three-band mapping for FCE (1–18/19–30/31–40) may blur top-end distinctions. As future work we will test finer buckets (e.g., four–five bands) and explore ordinal-aware conformal scores.

Finally, from a practical standpoint, even 4-bit LoRA fine-tuning of an 8B-parameter model requires a high-end GPU; institutions with modest hardware may still prefer smaller models. While calibrated prediction sets indicate \emph{how sure} the model is, they do not explain \emph{why} a script is low, medium or high; integrating rubric-aligned rationales is an important next step toward truly interpretable AES.

\bibliographystyle{acl_natbib}
\bibliography{anthology,custom} 

\end{document}